\def\year{2021}\relax
\documentclass[letterpaper]{article} 
\usepackage{aaai21}  
\usepackage{times}  
\usepackage{helvet} 
\usepackage{courier}  
\usepackage[hyphens]{url}  
\usepackage{graphicx} 
\usepackage{natbib}  
\usepackage{caption} 
\usepackage{amsfonts}
\usepackage{amsmath}
\usepackage{booktabs}
\usepackage{multirow}
\usepackage{xcolor}
\usepackage[position=bottom]{subfig}

\title{SML: Semantic Meta-learning for Few-shot Semantic Segmentation }
\author{
    Ayyappa Kumar Pambala\textsuperscript{\rm 1},
    Titir Dutta\textsuperscript{\rm 1},
    Soma Biswas\textsuperscript{\rm 1}
}
\affiliations{

    \textsuperscript{\rm 1} Electrical Engineering, Indian Institute of Science, Bangalore\\
    \{ayyappap, titird, somabiswas\}@iisc.ac.in 
}
\begin{document}
\maketitle
\begin{abstract}
The significant amount of training data required for training Convolutional Neural Networks has become a bottleneck for applications like semantic segmentation.
Few-shot semantic segmentation algorithms address this problem, with an aim to achieve good performance in the low-data regime, with few annotated training images.
Recently, approaches based on class-prototypes computed from available training data have achieved immense success for this task.
In this work, we propose a novel meta-learning framework, Semantic Meta-Learning (SML) which incorporates class level semantic descriptions in the generated prototypes for this problem.
In addition, we propose to use the well established technique, ridge regression, to not only bring in the class-level semantic information, but also to effectively utilise the information available from multiple images present in the training data for prototype computation.
This has a simple closed-form solution, and thus can be implemented easily and efficiently. 
Extensive experiments on the benchmark PASCAL-5i dataset under different experimental settings show the effectiveness of the proposed framework.
\end{abstract}

\section{Introduction}
Image segmentation is one of the fundamental problems in the field of computer vision. 
Traditional supervised segmentation methods~\cite{long2015fully, badrinarayanan2017segnet, lin2017refinenet} give impressive  results when large amounts of annotated data are available. 
However, this requirement of labeled training data for segmentation task is quite difficult, since annotating each and every pixel for huge amount of image data is highly expensive and cumbersome.
On the other hand, humans can identify any novel concept very easily even with very few examples of the same.
Few-shot semantic segmentation~\cite{wang2019panet, Zhang_2019_CVPR, siam2019adaptive} tries to address this problem, by working in the very low data regime, utilizing few annotated images from each class.

Meta-learning or learning-to-learn approaches have achieved very good performance for the problem of few shot learning~\cite{vinyals2016matching, snell2017prototypical}, and also for the segmentation application~\cite{dong2018few, rakelly2018conditional}.
Training meta-learning algorithms constitutes two stages of learning: 
(1) base-learner, which learns to predict an individual task at the episode level and 
(2) meta-learner, which learns to generalize by learning across a large number of training tasks/episodes. 
Significant amount of research has been done along these lines, but recently, the approaches based on computing class representatives or class prototypes have been very successful~\cite{wang2019panet,tian2020differentiable}.

In this work, we propose a novel meta-learning framework, termed as Semantic Meta-Learning or {\bf SML}, to address the few-shot semantic segmentation task by utilizing class-specific semantic information.
The proposed SML approach is also based on computing the prototypes corresponding to each class in the training data.
But, in contrast to other prototype-based meta-learning approaches in literature~\cite{wang2019panet, dong2018few}, SML does not compute the class prototypes as the average representation of all the visual feature embeddings.
Instead, they are learnt by incorporating the semantic knowledge of the particular class (obtained automatically from the class names) into the visual information obtained from the images by the base-learner.
In addition, we propose to utilise the visual feature embeddings obtained from multiple training images of the same class individually while computing the class prototypes, instead of averaging them.
Both these steps effectively bring the visual embeddings of the same class images closer to one another, and also maintain semantically meaningful intra-class distances between the class prototypes, even when few training images per class are available.
Inspired by the seminal work in~\cite{bertinetto2018meta}, these objectives are achieved through learning a linear function between the visual feature embeddings and the  semantic information or attributes using standard ridge regression method. 
Thus, in the proposed SML framework, this linear function has a closed-form solution to compute the prototypes, which makes the computation very efficient.
Extensive experiments on the benchmark PASCAL-5i dataset with different experimental settings show that the proposed SML framework is effective for few-shot semantic segmentation task.
SML also performs favorably with respect to the state-of-the-art, even for weaker annotations.
Thus, the contributions of this work are as follows:
\begin{enumerate}
    \item We propose a novel end-to-end meta-learing framework, SML, which can effectively integrate the attribute information and the visual feature embeddings for the meta-learner to generalize to new classes during testing.
    \item SML also utilises multiple images in the training data more effectively for computing better class-prototypes.
    \item Extensive evaluation on the PASCAL-5i shows that the proposed SML framework performs better or comparable to the state-of-the-art.
\end{enumerate}
The rest of the paper is organized as follows. 
The relevant literature is described in Section~\ref{relatedWork}.
The problem definition and the proposed approach are described in details in Section~\ref{problem} and Section~\ref{sml} respectively.
The experimental details are described in Sections~\ref{results} and the paper concludes with a brief summary in Section~\ref{conclusion}.

\begin{figure*}[ht!]
\begin{center}
\includegraphics[width=0.9\linewidth,keepaspectratio]{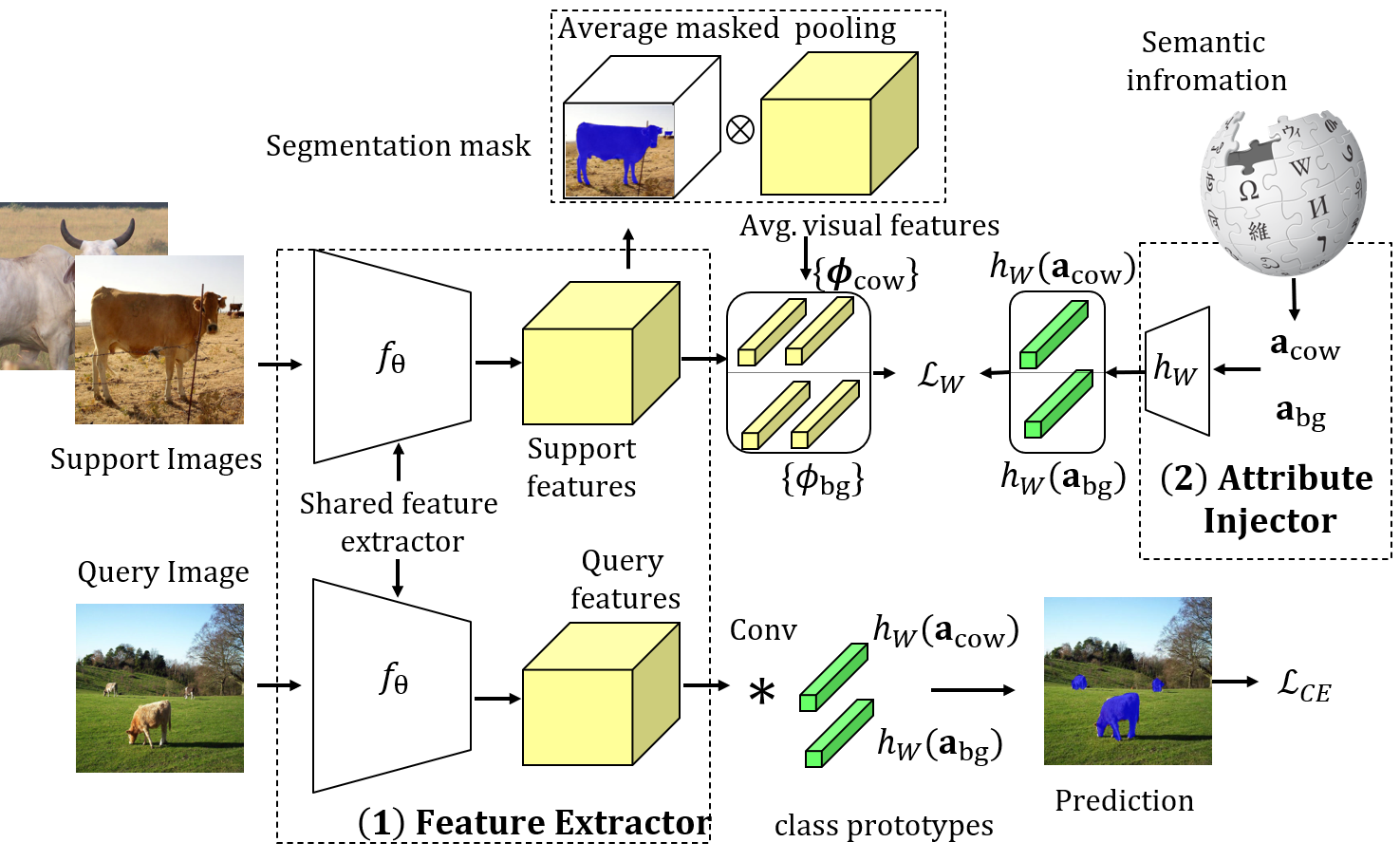}
\end{center}
\caption{Illustration of the proposed Semantic Meta-Learning (SML) framework for few-shot semantic segmentation. 
This illustration is for a single episode for 1-way 2-shot segmentation task. 
The base-learner learns the class-prototypes by integrating the visual information from the image (using average-masked features ${\phi}$) and the attribute information $\mathbf{a}$, which is accomplished using a linear function $h_W$.
These class-prototypes are used to obtain the final prediction over the query set. }
\label{figure_2}
\end{figure*}

\section{Related work}
\label{relatedWork}
Few-shot learning is a very active area of research in the field of computer vision, and several approaches have been proposed.
In this section, we give pointers to the relevant meta-learning based approaches in literature which addresses few-shot classification and semantic segmentation.
\subsubsection{Few-shot learning: }
Meta-learning based few-shot learning approaches can be broadly divided into (a) {\em metric learning} based approaches~\cite{vinyals2016matching, snell2017prototypical}, (b) {\em optimization} based approaches~\cite{finn2017model, ravi2016optimization, bertinetto2018meta}, etc. 
The goal of these approaches is to  leverage the support set images to learn to classify the images in the query set.
Matching network~\cite{vinyals2016matching} learns using soft weighted nearest neighbour scores obtained from support images.
Prototypical network~\cite{snell2017prototypical} evaluates support class means as the class-prototypes and the classification is performed on the query examples using Euclidean distance measure.
Relation network~\cite{sung2018learning} learns to find relations between small number of images at each episode, which contains both support and query images.
Recently, in differentiable solver~\cite{bertinetto2018meta}, the network is learned using ridge regression.
Though the proposed method is inspired from \cite{bertinetto2018meta}, we propose to additionally fuse the semantic information of the classes for the goal of  few-shot semantic segmentation.

\subsubsection{Few-shot segmentation: } 
Some of the early approaches in the literature for few-shot segmentation, follow the strategy in~\cite{vinyals2016matching, snell2017prototypical} to learn network parameters from the support images and perform pixel wise classification task on the query images.
Co-FCN~\cite{rakelly2018conditional} learns to segment the query images by fusing feature information of support images.
A metric learning based prototype learning is used in~\cite{dong2018few} for few-shot segmentation.
Weight imprinting mechanism of new classes using adaptive masked proxies is used in~\cite{siam2019adaptive}.
CANet~\cite{Zhang_2019_CVPR} uses iterative optimization solution and MetaSegNet~\cite{tian2020differentiable} uses a differentiable optimization solver for the segmentation task.
Recently, prototypical networks \cite{snell2017prototypical} for few-shot classification is adapted in \cite{dong2018few, wang2019panet} to perform the segmentation task.
PANet~\cite{wang2019panet} uses masked average pooling to obtain the prototypes, and also proposes a reverse alignment regularizer to learn the prototypes using a swapped-setting of support and query images. 
PPNet~\cite{liu2020part} uses multiple class prototypes with unlabelled data for the segmentation task.

Our method is inspired from \cite{wang2019panet} to obtain better class prototypes by incorporating the semantic class-descriptions (using the class names), into the base-learners to learn the segmentation task.
In essence, the base-learner exploits both visual information and semantic/attribute knowledge to obtain better prototypes and also to efficiently cluster the features of images of the same class.
In the next section, we explain the proposed approach.

\section{Problem Definition}
\label{problem}
Here, we discuss in details the proposed SML framework for the task of few-shot semantic segmentation.
First, we describe the notations used and the problem statement.
Let the annotated training data be denoted as $\mathcal{D}_{base} = \{\mathbf{I}, \mathbf{M}\}$, where, $\mathbf{M}$ represents the ground truth mask corresponding to the input image $\mathbf{I}$.
Let $\mathcal{C}_{base}$ be the set of base classes for which the annotated data is available. 
During testing, the goal is to segment images from a set of novel classes $\mathcal{C}_{novel}$, given only few images from each class. 
Let the number of images available per class during testing be denoted by $K$, and $K$ is usually less than or equal to $5$.
The testing classes do not overlap with the training ones, i.e. $\mathcal{C}_{base} \cap \mathcal{C}_{novel} = \phi$.

To address this task, a meta-learning based approach is adopted, where we attempt to imitate the testing scenario as closely as possible during training.
The entire training process is split into sub-tasks or episodes, referred to as episode-based training~\cite{vinyals2016matching, snell2017prototypical}.
Each task or episode $\mathcal{E} = (\mathcal{S}, \mathcal{Q})$ consists of a support set $\mathcal{S}$ and a query set $\mathcal{Q}$.
To emulate the testing scenario, each episode is constructed to be a $C$-way, $K$-shot segmentation task, i.e., given a support set of $K$ images and their ground truth masks from each of $C$ classes, the goal is to segment the images in the query set containing objects from one of these $C$ classes.
Thus, $\mathcal{S}$ contains randomly selected $C$ classes from the set $\mathcal{C}_{base}$ with $K$ images from each class, i.e., 
$\mathcal{S} = \bigcup \limits_{c=1}^{C} \{\mathbf{I}^s_{i,c}, \mathbf{M}^s_{i,c} \}_{i=1}^{K}$.
Similarly, the query set $\mathcal{Q} = \{\mathbf{I}_{j}^q, \mathbf{M}^q_{j}\}_{j=1}^{N_q}$ contains $N_q$ number of query images from the classes present in its support set.
The set of images in the support and query sets in every episode is strictly non-overlapping.
The network is trained by sampling several episodes in a meta-learning fashion.

In this work, segmentation is performed by computing semantically meaningful prototypes for each class. 
Given the class names present in the support (or query) set, any existing pre-trained language model, such as FastText~\cite{joulin2016fasttext} or Word2Vec~\cite{mikolov2013distributed} can be utilised to obtain the semantic or attribute vectors automatically corresponding to those object classes.
Let $\mathbf{a}_{c} \in \mathbb{R}^{d_a}$ denote the attribute vector corresponding to the $c^{th}$ class.
All image pixels in the support and query set which belong to the $c^{th}$ class will have the same semantic representation.
Note that, this information can be easily obtained given just the class names in $\mathcal{D}_{base}$, and does not require any additional information.
Attribute information is widely used in applications like ZSL~\cite{akata2015label}, but it has been relatively less explored in the context of few-shot learning.
\subsection{Main Idea of SML framework} 
An illustration of the SML framework for few-shot segmentation is shown in Figure~\ref{figure_2}.
The proposed framework is learnt in a meta-learning fashion consisting of several episodes, and Figure~\ref{figure_2} illustrates the base learner for $1$-way, $2$-shot scenario.
It has two main modules:
1) feature-extractor module and 
2) attribute-injector module.
For each episode, given the support images, their visual features are extracted using the feature extractor network. 
Their masks are used to obtain the foreground and background regions, which are in turn used to compute the average foreground ($\phi_{cow}$) and background ($\phi_{bg}$) feature embeddings for each support image by average mask pooling.
The class names in the support set is used by the attribute injector module to incorporate the semantic information into the base learner.
Specifically, this is achieved by using a linear function ($h_W$) to effectively combine the information from the attribute vectors and the visual features, towards the goal to obtain better class prototypes ($h_W(.)$).
These final prototypes of the class ($h_W({\bf a}_{cow})$) and the background ($h_W({\bf a}_{bg})$) are used to perform segmentation on the images in the query set.
The loss computed between the query prediction and its ground truth mask is used to update the base-learner, which constitute the feature extractor and the linear function $h_W$.
The meta-learner is learnt over several training episodes to generalize to segment novel classes with few examples per class.
We will discuss these modules in details in the next section.

The main difference between the proposed SML framework and other prototype-based few-shot segmentation approaches~\cite{wang2019panet, dong2018few} is the use of attribute information.
We feel that semantic information is extremely beneficial, specially in the low-data regime.
Instead of relying solely on the visual information from the images, SML effectively combines it with the semantic information for obtaining improved class prototypes.
In addition, the features from multiple support images are not averaged, rather they individually contribute to the prototype computation. 
This helps to bring the features from the same class closer to one another, thus effectively utilising all the additional support images.
The closed form solution of the linear function makes the learning very efficient.
Extensive experiments and analysis have been presented in Section~\ref{results}, which justifies the effectiveness of this framework.
Now, we will discuss the design of loss functions and meta-learning based training methodology followed in our work.

\section{Proposed SML Framework}
\label{sml}
In this section, we will describe the different components of the proposed SML framework.
The proposed meta-learning framework learns over a collection of episodes, and thus gets trained with data of all classes ($\mathcal{C}_{base}$) in the training set.
On the other hand, the \emph{base learner} learns over a single episode 
with data from randomly selected $C$-classes.
The base learner consists of a feature extractor module and an attribute injector module as described below.
\subsubsection{1. Feature-extractor module: }
Given an input image $\mathbf{I}$ from support or query, the feature extraction module $f_{\theta}$ extracts its 3-d feature representation $f_{\theta}(\mathbf{I}) \in \mathbb{R}^{h \times w \times d}$.
Here, $f_{\theta}(\mathbf{I})$ is the upsampled feature map and has the same height ($h$) and width ($w$) as the
original image and and $d$ denotes the number of channels of the final convolutional layer.
The set of learnable parameters in the feature extractor module is denoted as $\theta$.
This 3-d visual feature representation of an image can be considered as the $d$-dimensional embedding of each of  the image pixels, which either belong to the class of interest or the background.


As per the standard meta-learning set-up~\cite{vilalta2002perspective}, in each episode, the base learner first processes the data in the support set $\mathcal{S}$, and then computes the losses on the query set $\mathcal{Q}$.
Given the support set $\mathcal{S} = \bigcup \limits_{c=1}^{C} \{\mathbf{I}^s_{i,c}, \mathbf{M}^s_{i,c} \}_{i=1}^{K}$, for every image, its 3-d feature representation $f_{\theta}({\mathbf{I}_{i,c}^s})$
is computed using the feature extractor module.
This obviously contains the feature representation of both the object~(or, the foreground) and the background in the image.
The foreground and background features can be separated using the mask $\mathbf{M}_{i,c}^s$ provided with the input image.
As in several other works~\cite{siam2019adaptive, wang2019panet}, we also use average mask-pooling operation to obtain the foreground~(or, background) embedding as follows:
\begin{displaymath}
{\phi}_{i,c}^{s} = \frac{1}{|\mathcal{X}|} \sum_{\mathcal{X}}{f_{\theta}(\mathbf{I}_{i,c}^s) \odot \mathbf{1}_{(\mathbf{M}_{i,c}^s==c)}} \text{, for foreground} 
\end{displaymath}
\begin{equation}
\label{phi_embedding}
    {\phi}_{i,bg}^s = \frac{1}{|\mathcal{Y}|} \sum_{\mathcal{Y}}{f_{\theta}(\mathbf{I}_{i,c}^s) \odot \mathbf{1}_{(\mathbf{M}_{i,c}^s \neq c)}} \text{, for background}
\end{equation}
Here, $\odot$ represents the point-wise multiplication along the dimension of $d$.
$\mathcal{X}$ and $\mathcal{Y}$ denote the respective sets of spatial locations in the image $\mathbf{I}_{i,c}^s$ for which the corresponding indicator functions are activated. 
Thus, $\phi_{i,c}^s$ and $\phi_{i,bg}^s \in \mathbb{R}^{d}$ can be considered as the mask-pooling of foreground and background features separately in the visual embedding space evaluated as the mean over their spatial spread in the image.
We normalize the features,  $\phi_{i,c}^s = \frac{\phi_{i,c}^s}{||\phi_{i,c}^s||_2}$ and $\phi_{i,bg} = \frac{\phi_{i,bg}^s}{||\phi_{i,bg}^s||_2}$ in our experiments.
\subsubsection{2. Attribute-injector module: }
In some of the recent successful approaches~\cite{wang2019panet}, the image embeddings are solely utilised to obtain the class-representatives or prototypes.
We propose to augment this information with the semantic knowledge for better generalization to unseen classes with few examples during testing. 
Here, the semantic information is given by the attribute vectors, $\mathbf{a}_{c} \in \mathbb{R}^{d_a}$, which are the class-name embeddings for the $c^{th}$ class.
These attributes can be automatically extracted from a pre-trained \textit{FastText}  \cite{joulin2016fasttext} or \textit{Word2Vec}~\cite{mikolov2013distributed} language model.
Here, $d_a$ denotes the attribute dimension.
Given these semantic attributes of the classes in the training data, this module incorporates this information to compute better class prototypes for the segmentation task. 
In this work, we use a linear function for the attribute injector as given below: 
\begin{align}
    h_W(\mathbf{a}_{c}) &= \mathbf{W} \mathbf{a_{c}}
\end{align}
where, $\mathbf{W} \in \mathbb{R}^{d \times d_a}$ is the weight matrix containing all the trainable parameters.
Similarly, we also obtain the latent space representation of the background of the images as $h_W(\mathbf{a}_{bg}) = \mathbf{W}\mathbf{a}_{bg}$; where $\mathbf{a}_{bg}$ is the FastText or Word2Vec embedding of the word  \emph{background}, and thus is the same for all images.

In order to effectively incorporate the semantic information with the embeddings obtained from the images using equation~\eqref{phi_embedding} in the latent space,  we use the standard ridge regression~\cite{bertinetto2018meta, verma2017simple}. 
Thus, the loss function for learning $\mathbf{W}$ can be expressed as
\begin{align}
    \mathcal{L}_{W} &= || \Phi - \mathbf{W} \mathbf{A} ||_2^2 + \lambda ||\mathbf{W}||_2^2
\end{align}
where, 
$\Phi=\big[\{\phi_{i,c}^s|\phi_{i,bg}\}_{i=1}^K\big]_{c=1}^{C} \in \mathbb{R}^{d \times 2|\mathcal{S}|}$ 
and
$\mathbf{A} = \big[\{\mathbf{a}_{i,c}^s|\mathbf{a}_{i,bg}^s\}_{i=1}^{K}\big]_{c=1}^C \in \mathbb{R}^{d_a \times 2|\mathcal{S}|}$. 
$\lambda$ is a learnable hyper-parameter which is set experimentally to balance the L2-regularizer on the parameters $\mathbf{W}$.
The optimum set of parameters of the attribute injector module $\mathbf{W}$ is obtained by minimizing this loss function which has a closed-form solution given by
\begin{equation}
\label{closed_form_W}
    \mathbf{W} = {\Phi}\mathbf{A}^T(\mathbf{A}\mathbf{A}^T + \lambda \mathbf{I}_{d_a})^{-1}
\end{equation} 
where, $\mathbf{I}_{d_a}$ is an identity matrix of dimension $d_a$.

\subsection{Learning the base-learner parameters}
The goal is to leverage the query set images and its corresponding ground truth masks to learn the parameters of the feature extractor module and the linear function parameters $\mathbf{W}$, such that during testing, semantic segmentation can be performed using only few  images per class.
In each episode, given the support images with the corresponding masks from $C$-classes, $\Phi$ and $\mathbf{A}$ can be computed, from which the linear function parameters $\mathbf{W}$ are learnt as discussed before.
Note that, for each class $c$, the images corresponding to this class will have different feature representations ($\phi_{i,c}^s$) because of the intra-class variability, but they will all have the same semantic representation corresponding to the class name $\mathbf{a}_{c}$ for both support and query. 
This unique semantic representation of the $c^{th}$ class is considered its class prototype $h_W(\mathbf{a}_c) = \mathbf{W}\mathbf{a}_c$ in this work.

Now, given an image from the query set $\mathbf{I}_{j}^q$, we first compute its visual representation by passing it through the feature extractor module $f_\theta({\mathbf I}_{j}^q)$ and the feature corresponding to each pixel is normalized.
Next, the normalized feature representation $f_\theta(\mathbf{I_{j}^q})$ is convolved with the normalized class-prototypes for all the $C$-classes selected in that episode as follows:
\begin{align}
    \mathbf{S}_{j;c}^q &= f_{\theta}(\mathbf{I}_{j}^q) * h_W(\mathbf{a}_c), \text{ for }c = \{1,...,C\} \cup \{bg\}
\end{align} 
Here, $*$ stands for convolution operation and $h_W(\mathbf{a}_c)=\mathbf{W}\mathbf{a}_c$ is the class-prototype of the $c^{th}$ class.
We have used the same notations for the normalized representations for simplicity. 
Thus, $\mathbf{S}_{j;c}^q \in \mathbb{R}^{h \times w}$ denotes the 2-d mask of the query image with the class-similarities, which is finally used for classification.
Here, in addition to the foreground classes, similarity with the background class prototype \emph{bg} is also computed. 
To segment the query image, we perform pixel-wise classification of the query feature $\mathbf{S}_{j;c}^q$.
For this, we compute the logit-score for the pixel at location $(m,n)$ as the probability of the evaluated similarity feature $\mathbf{S}_{j;c}^q$ to belong to class $c$ as
\begin{align}
\label{eq:logits_pixel_wise}
    p(\mathbf{S}_{j;c}^q(m,n)) = \frac{exp(\alpha \mathbf{S}_{j;c}^{q}(m,n) + \beta)}{\sum\limits_{c \in {\{1,..,C\}\cup \{bg\} }}{exp(\alpha \mathbf{S}_{j;c} ^{q}(m,n)+\beta)}}   
\end{align}
where $\alpha, \beta$ are the scaling and bias parameters, respectively.
We learn the feature extractor module of the base learner by minimizing the cross-entropy loss function
\begin{align}
    \mathcal{L}_{CE} &= {\sum\limits_{(m,n) \in \mathbf{I}_{j}^q}{- \log p(\mathbf{S}_{j;c}^{q}(m,n))}}
\end{align}
Inspired by~\cite{wang2019panet}, we also interchange the samples in support $\mathcal{S}$ and query $\mathcal{Q}$ set to enhance the segmentation performance. 
Thus, we learn another set of parameters $\bar{\mathbf{W}}$ on the newly-constructed  $\mathcal{S}$ and  $\mathcal{Q}$ using~\eqref{closed_form_W}. 
Following similar steps as shown above, we compute the reverse alignment loss $\mathcal{L}_{R}$ as in~\cite{wang2019panet} which is given by,
\begin{equation}
    \mathcal{L}_{R} = {\sum\limits_{(m,n) \in \mathbf{I}^s}{- \log p({\mathbf{S}}_{c}^{s}(m,n)})}
\end{equation}
Note that subscripts are removed for the similarity predictions and for the images to avoid clutter.
The objective used to learn the feature extractor is given by
\begin{equation}
   \min_{\theta} \sum \limits_{(\mathcal{S}, \mathcal{Q})}\big( \mathcal{L}_{CE} + \mathcal{L}_{R} \big)
\end{equation}
This completes the learning of the  base-learner for a single episode.
To summarize, in each episode, first, the attribute injector module parameters $\mathbf{W}$ are learnt to generate class prototypes using the previously learnt base-learner, which are used to further fine-tune the feature extractor parameters $\theta$ by computing the pixel wise cross-entropy loss.
Using several such training episodes, the meta-learner learns to generalize and segment images containing novel objects during testing. 


\subsection{Prediction}
Once the meta-learner is trained using all the data over a number of training episodes, it can be used to perform segmentation on novel images in an episode, $\mathcal{E}= (\mathcal{S}, \mathcal{Q})$ which contains images from $C_{novel}$.
As in training, $\mathbf{W}$ computed using the visual image features of the support set and the attributes of the classes present in the testing set.
Using this, the class prototypes are obtained.
Finally, the prediction on the features extracted from the query image $\mathbf{I}_j^q \in \mathcal{Q}$ is performed as, 
\begin{equation}
    \hat{\mathbf{S}}_{j}^q(m, n) =  \underset{c \in \{1, .., C\} \cup\{bg\} }{\mathrm{arg \text{ } max}}~~ \alpha \mathbf{S}_{j, c}^q(m,n)+ \beta.
\end{equation}

\section{Experiments}
\label{results}
In this section, we describe the experiments performed to evaluate the effectiveness of the proposed SML framework.
\begin{table*}[ht!]
\footnotesize
\begin{tabular}{@{}l|c|ccccc|ccccc@{}}
\toprule
\multicolumn{1}{l|}{\multirow{2}{*}{\text{Method}}} & \multirow{2}{*}{Backbone}& \multicolumn{5}{c|}{1-shot}                                                    & \multicolumn{5}{c}{5-shot}                   \\ \cmidrule(l){3-12} 
\multicolumn{1}{l|}{}                    &    & split-0       & split-1       & split-2       & split-3       & Mean          & split-0 & split-1 & split-2 & split-3 & Mean \\ \midrule
\text{OSLM}~\cite{shaban2017one}            &   VGG16 & 33.6          & 55.3          & 40.9          & 33.5          & 40.8          & 35.9    & 58.1    & 42.7    & 39.1    & 43.9 \\
\text{co-FCN}~\cite{rakelly2018conditional} &   VGG16 & 36.7          & 50.6          & 44.9          & 32.4          & 41.1          & 37.5    & 50.0    & 44.1    & 33.9    & 33.9 \\
\text{SG-One}~\cite{zhang2020sg}             &    VGG16   & 40.2          & 58.4          & 48.4          & 38.4          & 46.3          & 41.9    & 58.6    & 48.6    & 39.4    & 47.1 \\
\text{AMP}~\cite{siam2019adaptive}       & VGG16   &     - &     -  &    -   & -      & 43.4      &      -   & -       & -       &  -        &  46.9
 \\
\text{PANet}~\cite{wang2019panet}          &       VGG16                      & 42.3          & 58.0          & 51.1          & 41.2          & 48.1          & 51.8    & 64.6    & 59.8    & 46.5    & 55.7 \\ 
\text{{\bf SML (Ours)}}                            &    VGG16 & {\bf 43.0}          & {\bf 59.0}          & {\bf 51.3}          & {\bf 41.4}          & {\bf 48.7}          & {\bf 52.3}    & {\bf 64.9}    & {\bf 61.0}    & {\bf 50.4}    & {\bf 57.1} \\ \midrule
\text{PANet}~\cite{wang2019panet}          & RN50 &44.0 & 57.5 &  50.8 &  44.0            &49.1         & 55.3&  67.2 & 61.2&  53.2&  59.2  \\ 
\text{PPNet}~\cite{liu2020part}          & RN50 & 47.8  & 58.7&   {\bf 53.8} &  45.6            &51.5          & \textbf{58.3} &   {\bf 67.8}  & \textbf{64.8}  &  \textbf{56.7}  & \textbf{61.9} \\
\text{CANet}~\cite{Zhang_2019_CVPR}          & RN50 &\textbf{52.5} & \textbf{65.9} &   51.3 & \textbf{ 51.9}           &\textbf{55.4}          & 55.5&   {\bf 67.8} & {51.9}  & 53.2 & 57.1 \\ 
\text{{\bf SML (Ours)}}                        &   RN50  & {47.4} & {59.7} &
\textbf{53.8} & {44.4} & {51.3} &    {56.0}       &    \textbf{67.8}     &  {62.1}	&     {54.0}  &   {60.0}    \\ \bottomrule
\end{tabular}
\caption{Performance evaluation of the proposed SML and comparison with the other state-of-the-art methods for both 1-way 1-shot and 1-way 5-shot experimental protocols on PASCAL-5i data. The results are reported in terms of mean-IoU.}
\label{table:main_result}
\end{table*}
\begin{table}[ht!]
\footnotesize
\centering
\begin{tabular}{@{}lccc@{}}
\toprule
Method                              & Backbone & 1-shot & 5-shot \\ \midrule
FG-BG~\cite{rakelly2018few}         & VGG16   & 55.0   & -      \\
Fine-Tuning~\cite{rakelly2018few}   & VGG16   & 55.1   & 55.6   \\
OSLSM~\cite{shaban2017one}          & VGG16   & 61.3   & 61.5   \\
co-FCN~\cite{rakelly2018conditional}& VGG16   & 60.1   & 60.2   \\
PL~\cite{dong2018few}               & VGG16   & 61.2   & 62.3   \\
A-MCG~\cite{hu2019attention}        & VGG16   & 61.2   & 62.2   \\
SG-One~\cite{zhang2020sg}           & VGG16   & 63.9   & 65.9   \\
AMP~\cite{siam2019adaptive}         & VGG16   & 62.2   & 63.8   \\
PANet~\cite{wang2019panet}          & VGG16   & 66.5   & 70.7   \\ 
{\bf SML (Ours)}                                 & VGG16   & {\bf 66.8}   & {\bf 71.0}    \\ \midrule

CANet~\cite{Zhang_2019_CVPR}        & RN50    & 66.2   & 69.6   \\
{\bf SML (Ours)}                                  & RN50    &\textbf{67.1}  &    \textbf{72.2 }   \\ \bottomrule
\end{tabular}
\caption{Evaluation~(binary mean-IoU) of segmentation results of proposed SML and comparison with other state-of-the-art for 1-way 1-shot protocol on PASCAL-5i data.}
\label{table:binaryIOU}
\end{table}

\subsubsection{Dataset Used:} 
We use PASCAL-5i~\cite{shaban2017one} dataset to evaluate the model performance. PASCAL-5i~\cite{shaban2017one} is derived from PASCAL VOC 2012~\cite{everingham2010pascal} and SBD \cite{hariharan2011semantic}. This dataset contains $20$ classes.
All the $20$-classes are divided into $4$ splits, with $5$-categories per split.
As is the standard practice, the proposed SML model is trained on $3$ splits~\cite{shaban2017one} and evaluated on the $4^{th}$ split. 

\subsubsection{Word-embeddings: }We use two semantic encodings in our work: (1) \textit{Word2vec}~\cite{mikolov2013distributed} is trained on Google News dataset~\cite{wang2018zero} which contains $3$-million words; (2) \textit{FastText}~\cite{joulin2016fasttext} is trained on Common-Crawl dataset~\cite{mikolov2017advances}.
We use these pre-trained models for extracting the class-name embeddings in our work.

\subsubsection{Evaluation metric: } Two widely-used metrices, Mean-IoU~\cite{shaban2017one, zhang2020sg} and Binay-IoU~\cite{rakelly2018conditional, dong2018few, hu2019attention} are used to report the segmentation performance of the proposed SML.
Mean-IoU calculates mean of Intersection-over-Union for all the foreground classes.
Binary-IoU calculates the Intersection-over-Union by treating all the foreground classes as one class and background class as one class. 

\subsubsection{Implementation Details:}
We implement our proposed approch using PyTorch \cite{paszke2017automatic}. 
We use TITAN-X $2040$ $12$GB GPU to run our experiments.
We use VGG-$16$~\cite{simonyan2014very}, and ResNet-$50$~\cite{he2016deep} models, pre-trained on ImageNet~\cite{russakovsky2015imagenet} as the feature extractor module for SML.  
In case of VGG-16, the image features are extracted from the output of $5^{th}$ convolution block.
For ResNet-$50$, the image features are extracted from the $4^{th}$ convolution block.
In our implementation, we modify ResNet-50 as follows: first two residual blocks use convolution with stride $1$, and last two blocks are designed with dialated convolutions with $2$, $4$, to increase the receptive field.
The number of episodes used to train the proposed SML is $30k$.
The SGD-solver with weight decay of $5 \times 10^{-4}$ and momentum of $0.9$ is used for the optimization.
We apply initial learning rate of $1 \times 10^{-3}$ and $1.75 \times 10^{-3}$ for VGG-$16$ and ResNet-$50$ respectively.
The learning rate is reduced by a factor of $0.1$ after every $10k$ iterations. 
The learnable parameter $\lambda$ in equation~\eqref{closed_form_W} is initialized to 100. 
We empirically set $\alpha = 10$ and $\beta = 1$ in equation~\eqref{eq:logits_pixel_wise}.
We have used pre-trained \textit{Word2Vec} model to extract the semantic embedding from class-names in our results and only used \textit{FastText} for additional analysis.
The proposed SML framework is evaluated over $1k$ testing episodes. 
To mitigate the sensitivity of the model to random initialization, we repeat the experiments $5$ times and report the mean-IoU. 
\begin{figure}[ht!]
 {\includegraphics[width=0.24\linewidth,keepaspectratio]{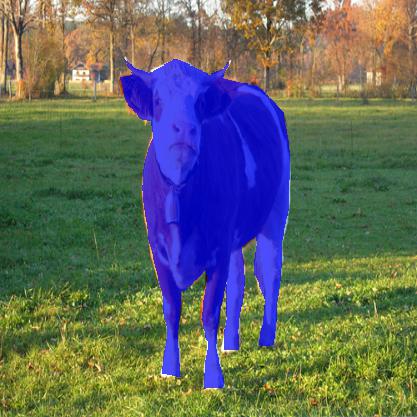}}
 {\includegraphics[width=0.24\linewidth,keepaspectratio]{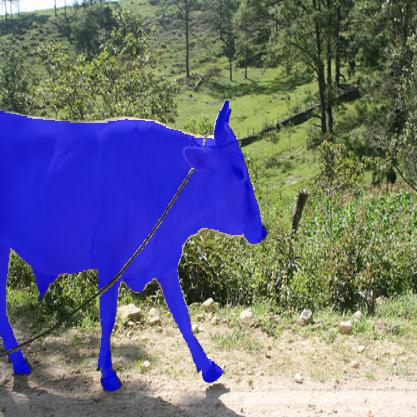}}
 {\includegraphics[width=0.24\linewidth,keepaspectratio]{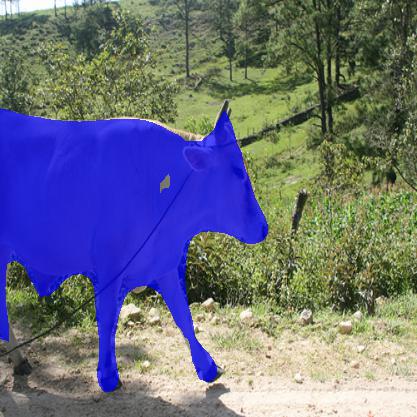}}
 {\includegraphics[width=0.24\linewidth,keepaspectratio]{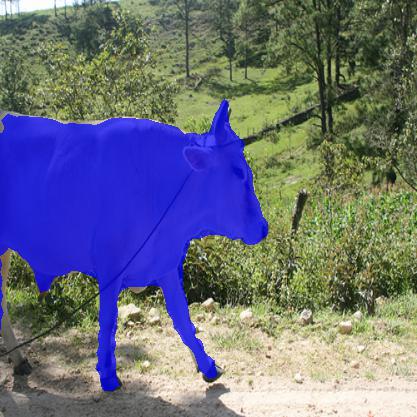}}
 \vspace{3mm}\\
  {\includegraphics[width=0.24\linewidth,keepaspectratio]{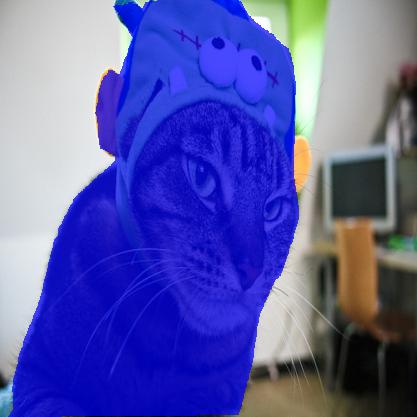}}
 {\includegraphics[width=0.24\linewidth,keepaspectratio]{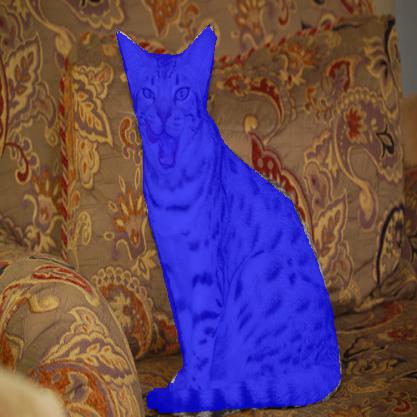}}
 {\includegraphics[width=0.24\linewidth,keepaspectratio]{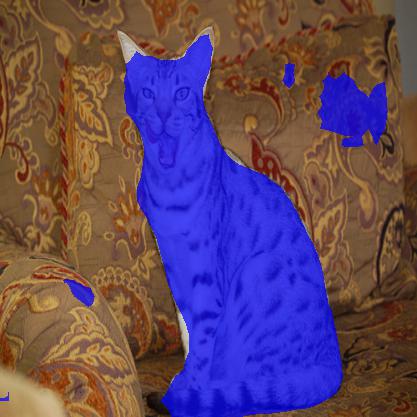}}
 {\includegraphics[width=0.24\linewidth,keepaspectratio]{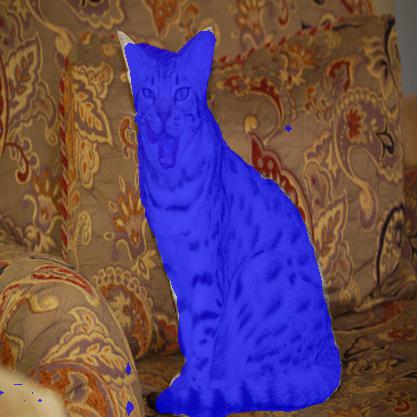}}
 \vspace{3mm}\\
  {\includegraphics[width=0.24\linewidth,keepaspectratio]{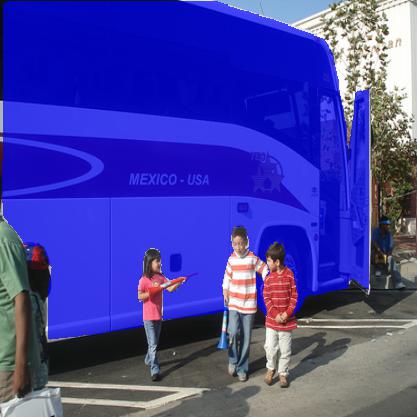}}
 {\includegraphics[width=0.24\linewidth,keepaspectratio]{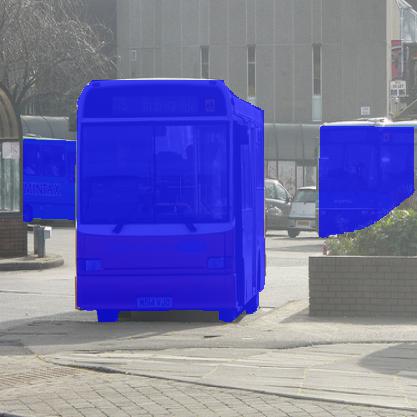}}
 {\includegraphics[width=0.24\linewidth,keepaspectratio]{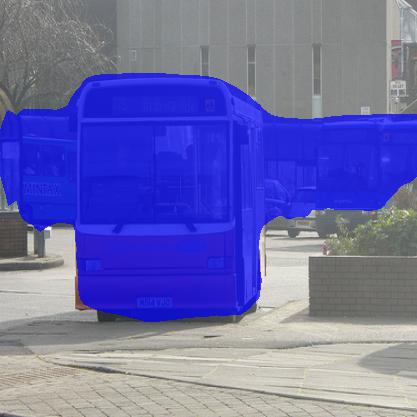}}
 {\includegraphics[width=0.24\linewidth,keepaspectratio]{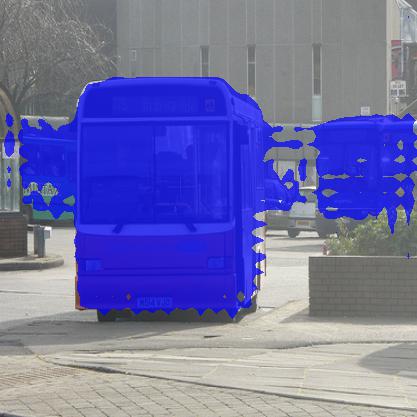}}
 \vspace{0mm}\\
 \subfloat[][Support]{\includegraphics[width=0.24\linewidth,keepaspectratio]{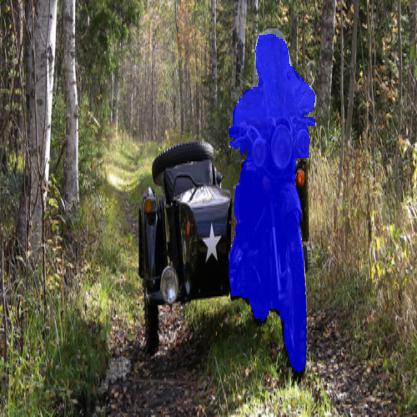}}\hspace{0.1mm}
 \subfloat[][GT]{\includegraphics[width=0.24\linewidth,keepaspectratio]{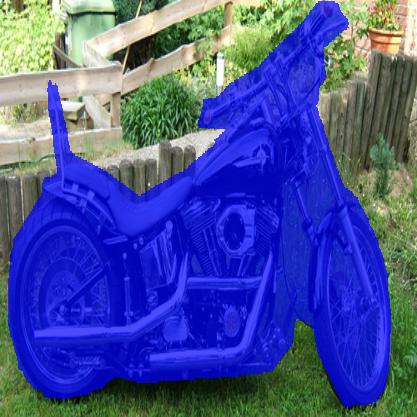}}\hspace{0.1mm}
 \subfloat[][VGG-16]{\includegraphics[width=0.24\linewidth,keepaspectratio]{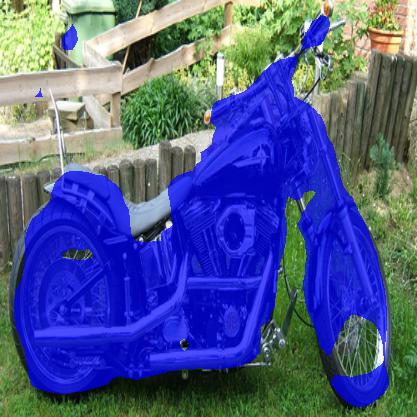}}\hspace{0.01mm}
 \subfloat[][ResNet-50]{\includegraphics[width=0.24\linewidth,keepaspectratio]{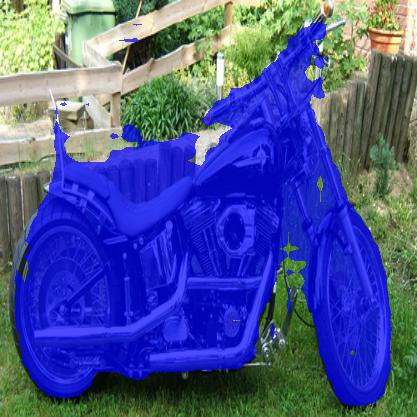}}
 \caption{1-way 1-shot segmentation results using proposed SML - (a) Support set image; (b) GT: query image with ground-truth mask; (c) and (d) are  segmentation results using VGG-16 and ResNet-50 as the feature extractors. Best viewed in color and when zoomed.}
 \label{fig:analysis}
\end{figure}
\begin{figure*}[t!]
 \centering
  {\includegraphics[width=1\linewidth,keepaspectratio]{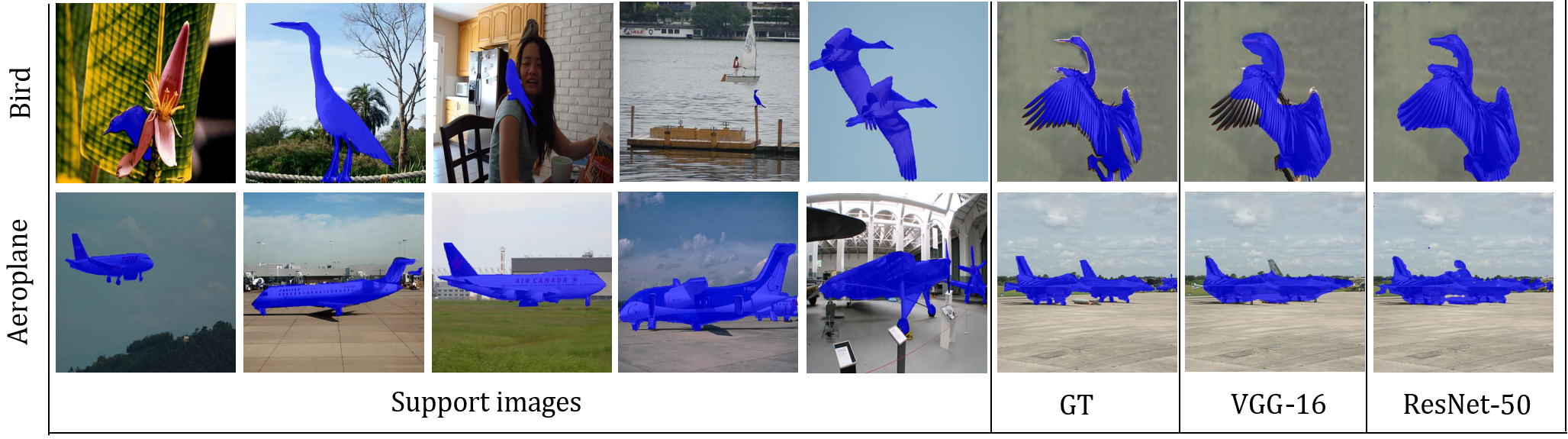}}\\
 \caption{1-way 5-shot segmentation results using proposed SML. First five columns shows the support set images with ground-truth masks. The remaining columns show the query image with ground-truth mask and segmentation results using VGG-16 and ResNet-50 as the feature extractors respectively.  Best viewed in color and when zoomed.}
  \label{figure:1way5shot}
\end{figure*}
\vspace{-0.2cm}
\begin{table}
\footnotesize
\centering
\begin{tabular}{@{}lccccc@{}}
\toprule
Method & split-0 & split-1 & split-2 & split-3 & Mean  \\ \midrule
PANet (VGG-16) & -       & -       & -       & -       & 45.1  \\ %
{\bf SML} (VGG-16)  & 40.5& 54.8  & 47.2    &39.5    & 	{\bf 45.5}\\ \midrule
{\bf SML} (ResNet-50)    & 43.2 &	55.6 &	49.5	&44.4 &	\textbf{48.1} \\ \bottomrule
\end{tabular}
\caption{Mean-IoU performance of proposed SML for 2-way 1-shot segmentation task.}
\label{table:2way1shot}
\end{table}

\subsection{Comparison with state-of-the-art methods }
We compare the proposed SML with several recent state-of-the-art methods - CANet~\cite{Zhang_2019_CVPR}, PANet~\cite{wang2019panet}, PPNet~\cite{liu2020part}, AMP~\cite{siam2019adaptive}, SG-One~\cite{zhang2020sg} etc.
Here we discuss the results for different testing protocols.
\subsubsection{1-way 1-shot and 1-way 5-shot:} 
We present the results for 1-way 1-shot and 1-way 5-shot semantic segmentation in Table \ref{table:main_result} for both VGG-$16$ and ResNet-$50$ backbones as the feature extractor.
The mean-IoU performances reported in the table for all the other approaches are taken directly from \cite{wang2019panet, liu2020part}.
We observe that for VGG-16 backbone, the proposed SML outperforms all the other approaches for both the protocols. 
The performance improvement compared to the other approaches is significantly more for 5-shot protocol.
This implies that SML is able to effectively use all the images in the support set to improve the segmentation performance. 
For ResNet-$50$ backbone, SML performs at par with the recent PPNet~\cite{liu2020part}.
Though CANet~\cite{Zhang_2019_CVPR} performs better, it is evaluated with multi-scale input unlike the others, as also noted in~\cite{liu2020part}. 

We also compare the SML performance with the state-of-the-art methods using binary-IoU as the evaluation criteria.
These results are reported in Table \ref{table:binaryIOU}.
The results for all the other methods are taken directly from~\cite{wang2019panet, siam2019adaptive}.
We observe that SML outperforms all other approaches for both the backbones.
Specifically, for ResNet-$50$, it gives an improvement of $0.9\%$ and $2.6\%$ for 1-shot and 5-shot respectively.

We present the segmentation results for few images using SML in Figure~\ref{fig:analysis} and Figure~\ref{figure:1way5shot} for 1-way 1-shot and 1-way 5-shot experimental protocols respectively.
We observe that SML provides good segmentation results  even with significant background clutter. In general, ResNet-$50$ gives better results than VGG-$16$ backbone, which is also validated by the quantitative results. 
\subsubsection{2-way 1-shot: }
We further investigate the performance of SML for $2$-way $1$-shot protocol.
From Table~\ref{table:2way1shot}, we observe that SML performs favorably compared to PANet~\cite{wang2019panet} for the same VGG-$16$ backbone. 
\begin{table}[t!]
\footnotesize
\begin{tabular}{@{}l|ccc|ccc|@{}}
\toprule
\multicolumn{1}{c|}{\multirow{2}{*}{Method}} & \multicolumn{3}{c|}{1-shot}                                                           & \multicolumn{3}{c|}{5-shot}                                                           \\ \cmidrule(l){2-7} 
\multicolumn{1}{c|}{}                        & \multicolumn{1}{c|}{D} & \multicolumn{1}{c|}{BB} & \multicolumn{1}{c|}{S} & \multicolumn{1}{c|}{D} & \multicolumn{1}{c|}{BB} & \multicolumn{1}{c|}{S} \\ \midrule
PANet (VGG-16)                                      & 48.1                       & 45.1                    & 44.8                          & 55.7                       & 52.8                    & 54.6                          \\
{\bf SML} (VGG-16)                                         & 48.7                       & 45.5                    & 45.0                          & 57.1                       & 53.7                    & {\bf 55.8}                          \\ \midrule
{\bf SML} (ResNet-50)                                          & 51.3                       & 47.9                    & 49.2                          & 60.0                       & 55.1                    & {\bf 57.6}                          \\ \bottomrule
\end{tabular}\caption{Mean-IoU of proposed SML with different image-annotations. D: Dense, BB: Bounding box, S: Scribble.}
\label{tab_weak_label}
\end{table}
As expected, the results are significantly better with ResNet-$50$ backbone. \\ \\
{\bf Additional Analysis: } Here we further analyze the proposed SML framework to better understand it. \\ 
{\bf {\em 1. Evaluation with weaker annotations: }}
We experiment with weaker annotations of the images which are cheaper and easy to obtain~\cite{lin2016scribblesup}, instead of the pixel-by-pixel dense annotations, which is extremely expensive and time-consuming. 
We observe from Table~\ref{tab_weak_label} that even for weaker annotations the proposed SML performs better than the PANet architecture \cite{wang2019panet}. \\
{\bf {\em 2. Evaluation with different semantic embedding: }}
All experiments in this paper are reported with \textit{Word2Vec} as the embedding-framework. 
We also performed experiment with  \textit{FastText}~\cite{joulin2016fasttext}. Those results, using both embeddings for $1$-way $1$-shot and $1$-way $5$-shot protocols, are summarized in Table~\ref{table:ablation}.
We observe that though both perform well, \textit{Word2Vec} embeddings are slightly better compared to \textit{FastText} for our application.
\begin{table}[t!]
\centering
\begin{tabular}{@{}lcc@{}}
\toprule
Semantic Knowledge  & 1-shot & 5-shot \\ \midrule
Word2Vec        & 51.3   & 60.0   \\
FastText        & 50.9   & 59.5       \\ \bottomrule
\end{tabular}
\caption{Mean-IoU for 1-way 1-shot and 5-shot segmentation tasks with different semantic-embeddings.}
\label{table:ablation}
\end{table}

\section{Conclusion}
\label{conclusion}
In this paper we have proposed a novel Semantic Meta-Learning (SML) framework which utilises the semantic information of object classes for the task of few-shot semantic segmentation.
Towards that goal, we introduced a novel attribute-injector module in a traditional meta-learning setting.
We performed extensive experiments with different protocols and observed that the proposed framework performs similar or better as compared to the state-of-the-art.
{
\bibliography{citations}
}
\end{document}